\documentclass[journal,twoside]{IEEEtran}
\usepackage{times}
\usepackage[table,dvipsnames]{xcolor}
\makeatletter
\let\NAT@parse\undefined
\makeatother
\usepackage[square, numbers]{natbib}
\usepackage{multicol}
\usepackage[bookmarks=true]{hyperref}
\usepackage{pifont} %for \ding{41}

\IEEEoverridecommandlockouts                              % This command is only needed if 
\usepackage{graphics} % for pdf, bitmapped graphics files
\usepackage{epsfig} % for postscript graphics files
\usepackage{mathptmx} % assumes new font selection scheme installed
\usepackage{amsmath} % assumes amsmath package installed
\usepackage{amssymb}  % assumes amsmath package installed

\usepackage{color}

\usepackage{graphicx}
\usepackage{balance}
\usepackage{multirow}
\usepackage{caption}
\usepackage{enumerate}
\usepackage{wrapfig}
\hypersetup{
linkcolor=BrickRed
,citecolor=Green
,filecolor=Mulberry
,urlcolor=NavyBlue
,menucolor=BrickRed
,runcolor=Mulberry
,linkbordercolor=BrickRed
,citebordercolor=Green
,filebordercolor=Mulberry
,urlbordercolor=NavyBlue
,menubordercolor=BrickRed
,runbordercolor=Mulberry
}

\title{Vision-Force Admittance Learning for\\Peg Insertion into a Movable Hole\vspace{-0.3cm}
}
\author{Yuzhong Chen$^{1}$, Yongqing Liang$^{1}$, Yunzhi Xu$^{2}$, Irving Fang$^{1}$, Chase Kidder$^{2}$, Hui-ping Wang$^{2}$,\\ Raihan Haque$^{2}$, Yubiao Zhang$^{2}$, Chen Feng\textsuperscript{1,\ding{41}}\\
{\tt\small \url{https://ai4ce.github.io/VFAL/}}
\thanks{Manuscript received: June 6, 2026; Revised: August 14, 2026; Accepted: August 24, 2026.}
\thanks{This paper was recommended for publication by Editor Markus Vincze upon evaluation of the Associate Editor and Reviewers comments.}
\thanks{$^{1}$New York University, Brooklyn, NY 11201, USA}
\thanks{$^{2}$General Motors, Warren, MI 48073, USA}
\thanks{\ding{41} Corresponding author ({\tt\small \href{mailto:cfeng}{cfeng@nyu.edu}}). The authors gratefully acknowledge the collaboration, support, resources, and technical expertise provided by the General Motors Autonomous Robotics Center (ARC) and the Robotics Intelligence team, which were instrumental to this work.}
\thanks{Digital Object Identifier (DOI): see top of this page.}
}

\IEEEaftertitletext{
\begin{center}
    \vspace{-0.5cm}
    \centering
    \begin{tabular}{cc}
         \includegraphics[height=6.3cm]{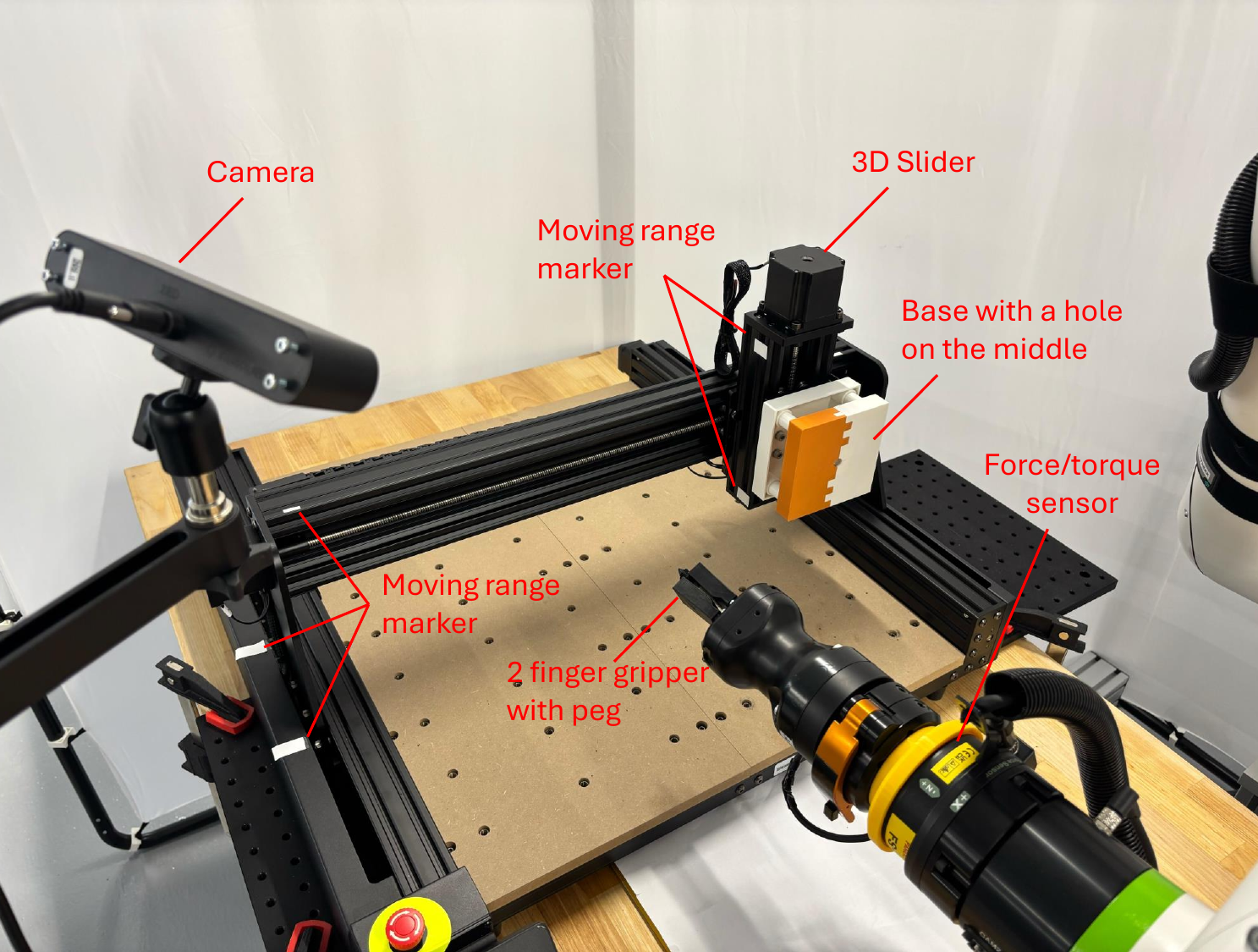}& 
         \includegraphics[height=6.3cm]{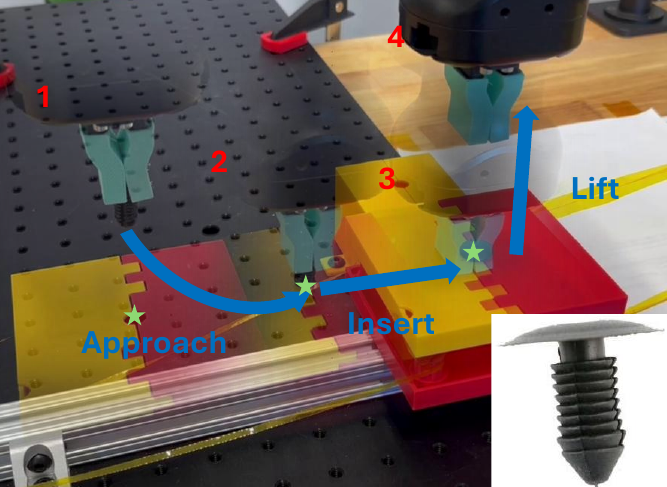} \\
         \text{(a) Experiment Setup} & \text{(b) Insertion Process}
    \end{tabular}
    \captionof{figure}{Overview of our real-world experiments. (a) shows the mobile base insertion setup. In (b), when the base moves with unknown motion, VFAL drives the robot to an approximate pre-insertion pose, then performs precise insertion using asynchronous force and visual feedback. After insertion, the robot lifts the end-effector.}
    \label{fig:fig1}
\end{center}%
}

\begin{document}

\markboth{IEEE Robotics and Automation Letters. Preprint Version. Accepted August, 2026}%
{Chen \MakeLowercase{\textit{et al.}}: Vision-Force Admittance Learning for Peg Insertion into a Movable Hole}
\maketitle

\maketitle

%%%%%%%%%%%%%%%%%%%%%%%%%%%%%%%%%%%%%%%%%%%%%%%%%%%%%%%%%%%%%%%%%%%%%%%%%%%%%%%%
\begin{abstract}

% abstract: research questionn + method + experiment + result
Precise manipulation in dynamic environments, whether induced by a mobile robot base or a target with unknown motion, remains a major challenge in robotics. Manipulation in dynamic environments introduces substantial uncertainty, which fundamentally conflicts with the tight precision requirement of precise tasks such as peg-in-the-hole. We propose a Vision-Force Admittance Learning (VFAL) framework that fuses asynchronous visual feedback with a high-frequency force-based model, using visual pose estimations as a regularization term. VFAL adapts insertion strategies online to dynamic motion while maintaining millimeter-level precision. To obtain robust, low-frequency pose information, we employ state-of-the-art vision foundation models for visual pose estimation. Additionally, we incorporate failure recovery mechanisms to enhance overall robustness. We validate our approach in real-world experiments, demonstrating high success rates and strong adaptability to various pegs and dynamic environments.
\end{abstract}

% \vspace{-0.15cm}

% \IEEEpeerreviewmaketitle

\section{Introduction}
% \vspace{-0.15cm}
% Industrial widely known task(no tactile, to more generalized force-based), challenge in the task(with current method), motivated by human, proposed method, contribution

% include: force based control can solve, not all robot can do that, reduce requirement of robot controller
%  while a force based control can achieve ..., it's not straightforward for robots with only position based control. most robot on the market don't have ...

% push nut insertion, fix base, moving target, no enough deformable peg in the hole, not too much about moving line
% in general
% application background
% The peg-in-hole problem is a widely studied assembly task in the robotic community. Most prior approaches focus on fixed-base configurations. However, extending peg-in-hole insertion to mobile or dynamic contexts, such as furniture assembly with a mobile robot or part installation on a production line, has received limited attention. The difficulty arises from the additional uncertainty introduced by robot or target motion, which significantly reduces precision. In this work, we specifically focus on the dynamic peg insertion problem, where a peg must be inserted into a moving base. This scenario amplifies the challenges of conventional peg-in-hole assembly by combining mobility, precision, and real-time adaptability.

Precise manipulation is a long-standing challenge in robotics and has been extensively studied in static settings. Most existing approaches assume fixed-base robots interacting with stationary targets, where uncertainty is primarily from unknown target pose and contact behavior, and high accuracy is attainable. In contrast, precise manipulation in dynamic environments, such as mobile manipulation or part installation on moving production lines, has received comparatively limited attention. The key difficulty arises from additional uncertainty induced by relative motion between the robot and the target, which greatly complicates accurate execution. In this work, we study precise manipulation in dynamic environments using peg-in-movable-hole as a representative task. This scenario amplifies the challenge by coupling motion with high accuracy requirements and the need for real-time adaptation.

% static, precise/dynamic?
Achieving success in such settings requires robot systems that combine high precision with strong adaptability to environmental changes. Prior work on precise manipulation has demonstrated high precision using impedance control~\cite{raibert1981hybrid,khatib1986motion}, compliant strategies~\cite{whitney1982quasi,park2020compliant}, and learning-based force regulation~\cite{beltran2020learning,luo2019reinforcement,tracy2025efficient}, enabling tasks such as peg-in-hole insertion and assembly~\cite{chen2025robust,negi2025learning}. However, these methods largely remain limited to static or quasi-static scenarios~\cite{whitney1982quasi,chhatpar2001search}. Conversely, dynamic manipulation techniques emphasize robustness to motion through visual servoing~\cite{liu2025visual}, prediction~\cite{uppal2024spin,xiong2024adaptive}, or reactive control~\cite{burgess2023architecture}, yet typically offer limited precision~\cite{burgess2023architecture,yang2024harmonic}.
% current method modality fusion side
% To succeed in such settings, robotic methods must achieve both high precision and strong adaptability to environmental changes. Vision-based approaches are highly robust to environmental variations~\cite{uppal2024spin} and generalize well across tasks~\cite{black2024pi0visionlanguageactionflowmodel}. However, they often struggle with latency, which reduces their precision in rapidly changing environments~\cite{luo2025precise}. In contrast, force-based methods are highly precise~\cite{beltran2020learning} and responsive~\cite{tracy2025efficient}, but suffer from issues caused by discontinuous contact dynamics~\cite{negi2025learning}. Recent efforts have explored combining vision and force-based methods to achieve both high precision and robustness~\cite{zhao2025touch, ichiwara2022contact}. These approaches either sacrifice speed due to differing input frequencies and processing complexities~\cite{ichiwara2022contact}, or fail to effectively utilize visual information during contact-rich manipulation phases~\cite{zhao2025touch}.
% one example, one drawback for each keypoint in related

Humans, by contrast, achieve precise manipulation in dynamic environments by flexibly integrating multiple sensing modalities: combining low-frequency, high-latency visual perception with high-frequency, low-latency force feedback to achieve both adaptability and precision~\cite{hutmacher2019there,helbig2007optimal}. This motivates the development of robotic methods capable of fusing asynchronous modalities in a complementary manner.

% not enough force control robot -> why not impedance instead -> develop based on position control/benefit of position control to force
A further practical challenge lies in hardware accessibility. Many robot arms lack direct force control and instead rely on position control. Implementing force control indirectly on such systems often introduces latency and instability~\cite{khatib1986motion}. Moreover, while pose is exploitable enough by vision-based models, force information is harder to incorporate into visual methods. These factors highlight the need for methods that can effectively utilize force feedback on position-controlled robots while still benefiting from visual priors.

% proposed method
To address these challenges, we present VFAL, a multimodal manipulation framework that asynchronously fuses vision and force for dynamic, precise insertion. VFAL builds upon Linear Model Learning (LML)~\cite{tracy2025efficient} for efficient online adaptation during insertion. Specifically, VFAL comprises two key components: (i) a vision-based pose estimation module that provides robust, CAD-based pose estimates at low frequency, and (ii) a force-based model with asynchronous visual referencing, in which model parameters are updated online using control and force feedback starting from a bootstrap initialization. Force-based predictions are regularized using the asynchronously updated pose. To improve robustness, VFAL incorporates recovery strategies for common failure modes.

% proposed contribution(system by demo, modality fusion by abalation, robot demo)
We claim these contributions:
% make it short, at most around 3 lines each
\begin{enumerate}
    \item We introduce a novel multimodal insertion framework that asynchronously fuses force and vision to enhance precision and adaptability in peg-in-movable-hole tasks.
    % define in commonsense(dynamic, precise, deformable)
    % improved capibality
    % tech highlight
    % \yq{I think it is not concise. We propose a robust manipulation scheme with xxx regularization. It overcomes the traditional methods failures on precise insertion tasks ..}
    \item We propose an asynchronous modality fusion pipeline that leverages predicted target pose as a regularization term for high-frequency admittance-based control.
    % new method
    \item We deploy VFAL on a position controlled robot and demonstrate robust peg insertion into a moving base.
    % experiment
\end{enumerate}

% \vspace{-0.15cm}

\section{Related Works}
\begin{figure*}
    \centering
    \includegraphics[width=0.9\textwidth]{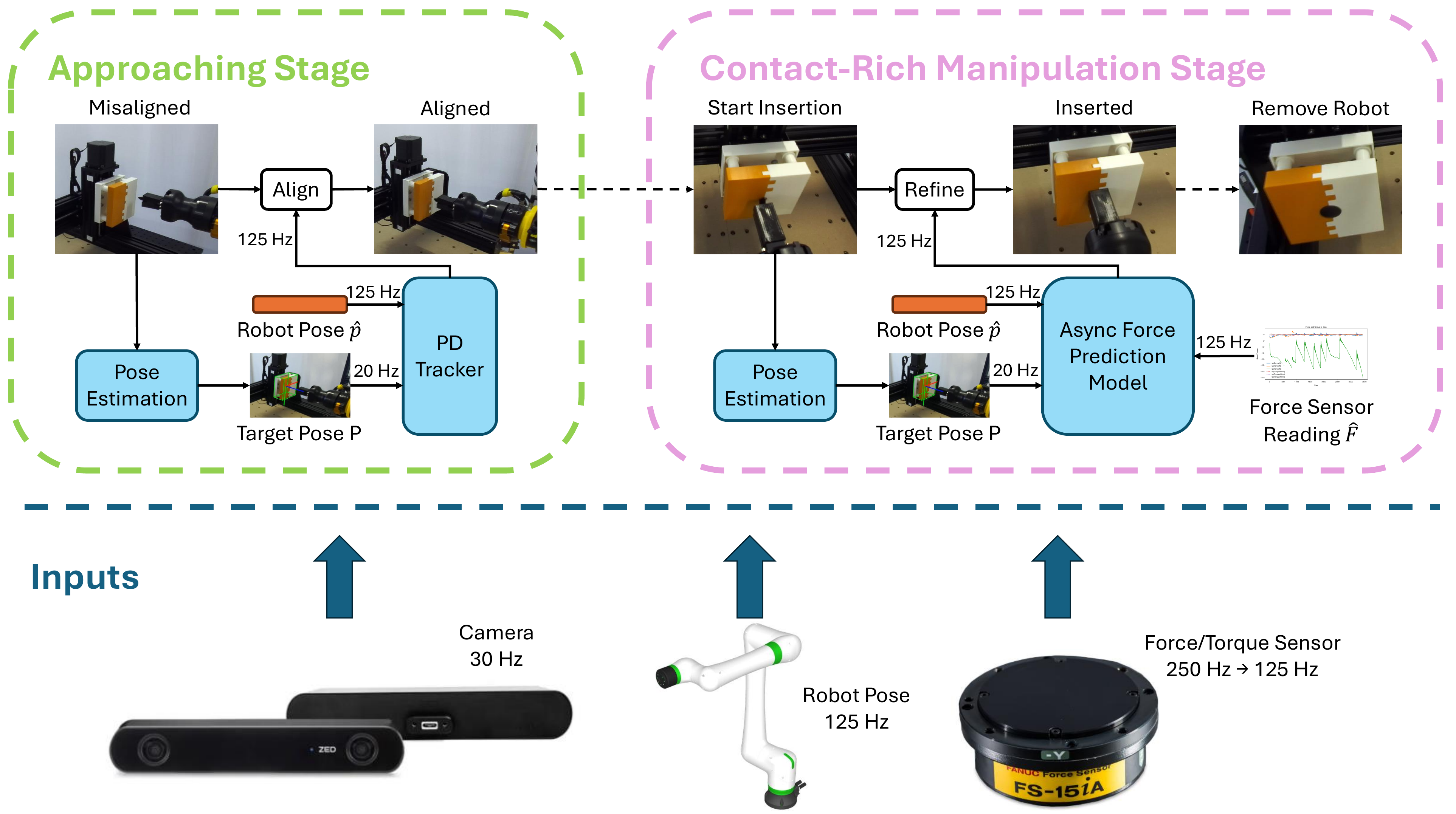}
    \caption{\textbf{Algorithm Workflow.} A two-stage pipeline is designed for the manipulation process. In the first stage, before making contact with the base, the robot approaches the target using the pose estimation result. In the second stage, once contact is detected via the force sensor, the model predicts the optimal next step control based on the current state and force readings, while referencing the asynchronously updated pose estimation as a regularization term. }
    \label{fig:workflow}
    \vspace{-0.5cm}
\end{figure*}

% \vspace{-0.15cm}
% peg in hole?
\subsection{Peg-in-Hole Methods}
% \vspace{-0.1cm}
The peg-in-hole problem has been extensively studied and remains popular in the robotics community due to its high precision requirements and widespread use in real-world applications. Early work primarily relied on model-based methods, including hybrid force/position control~\cite{whitney1982quasi, raibert1981hybrid} and analytical contact modeling~\cite{chen2025robust, chhatpar2001search}. To address large initial misalignment, search-based~\cite{park2020compliant, chhatpar2001search} strategies have been employed to locate holes initially outside the effective force-sensing region. More recently, learning-based approaches have gained attention for their strong generalization capability~\cite{chenlearning, inoue2017deep} and for incorporating multi-modal sensing~\cite{chenlearning, tao2025peg}. Beyond rigid peg-in-hole settings, some studies have also explored assembly with deformable objects~\cite{wu2025robotic, luo2018deep}.

However, previous methods mainly assume a static environment~\cite{park2020compliant, chen2025robust, tao2025peg}, where the hole is fixed but its position may be initially unknown. In contrast, our research focuses on a dynamic environment in which the hole position may change over time. This setting requires a model capable of responding to environmental changes while maintaining global sensing capability to handle large initial misalignment.

% \subsection{Mobile Manipulation Methods}
% % general introduction for mobile manipulation
% % Scene limitation -> precision
% Mobile manipulation has been extensively studied for its ability to extend the reach of robots and provide improved viewpoints for perception and planning~\cite{jauhri2024active, liu2025visual, yang2024harmonic}. Considerable progress has been made in household settings, where generalizable frameworks demonstrate robust performance across diverse tasks~\cite{qiu2024learning, gu2022multi, xiong2024adaptive}. However, research focusing on \emph{high-precision mobile manipulation} remains limited.

% Most approaches assume a static environment during execution~\cite{fu2024mobile, qiu2024learning}, which simplifies planning but fails to capture the complexity of real-world settings. Recent mobile rearrangement work~\cite{gumulti} introduces the challenge that manipulation itself alters the environment. Others attempt to anticipate object movement when planning grasps or interactions~\cite{burgess2023architecture}. Despite these advances, the induced changes are often insufficiently dynamic and fail to meet the precision demands in real-world precise assembly tasks. In particular, scenarios involving a moving target with high precision remain underexplored, yet are critical for bridging mobile manipulation with real-world high-precision assembly.

% methods and their draw backs: mobile parts have predictable behavior, not precise enough
%   moving base
%   moving target

% \vspace{-0.15cm}
\subsection{Force-Based Robot Manipulation}
% \vspace{-0.1cm}
% % general introduction for background for force based method
Force-based methods are well suited for precise manipulation due to their ability to operate at high frequency with low latency~\cite{inoue2017deep}, and their higher sensor precision in contact-rich manipulation compared to visual sensors. Tactile sensing provides rich contact information~\cite{matsubara2017active, zhao2025embedding}, but tactile sensors face durability challenges in industrial contexts, where long-term usage, chemical exposure, and temperature variation are common~\cite{zhao2025polytouchrobustmultimodaltactile}. Additionally, leveraging tactile data typically requires complex neural networks, whose computation latency limits their capability in high-frequency scenarios~\cite{calandra2018more}.

In contrast, force-torque sensors yield low-dimensional, high-frequency signals that are lightweight to process. Classical control methods such as Model Predictive Control (MPC) have been widely applied in this setting~\cite{kleff2022introducing, le2024fast}. However, MPC requires modeling of dynamics and lacks generalizability across varying tasks~\cite{katayama2023model}. Learning-based force control offers an alternative: these methods generalize without explicit dynamics models~\cite{inoue2017deep, negi2025learning, luo2019reinforcement}. Yet large models often fail to fully exploit the fast feedback provided by force sensors, while smaller models struggle in complex scenarios\textemdash{}such as mobile peg-in-hole insertion with deformable fasteners\textemdash{}where variability challenges robustness.

Online learning presents a promising path for adapting to dynamic and uncertain conditions~\cite{zhang2023efficient, tracy2025efficient, le2021learning}. Nevertheless, adaptation during the early stages of contact is often unstable, leading to frequent failures~\cite{zhang2023efficient, tracy2025efficient}, or requiring ground-truth demonstrations to guide the model~\cite{le2021learning}, which are difficult to obtain for precise manipulation in dynamic environments. These limitations motivate the use of global, but potentially stale and imprecise information to assist local force-based adaptation in challenging mobile assembly settings.

\subsection{Vision-Force Fusion for Robotic Manipulation}
% \vspace{-0.1cm}
% % Pure vision based + vision/force fusion: slow responsing, not suitable to dynamic ennvironment

Vision provides rich global context and improves robustness to environmental variation~\cite{uppal2024spin, black2024pi0visionlanguageactionflowmodel, luo2025precise}. However, the high latency of visual processing hinders its effectiveness in dynamic and precise tasks where millisecond-level reactivity is preferred~\cite{black2024pi0visionlanguageactionflowmodel, li2023pvt++}.

Hybrid approaches have attempted to combine vision and force sensing to balance global context with precision~\cite{zhao2025touch, ichiwara2022contact, he2025foar}. While conceptually appealing, these methods often encounter practical mismatches: force sensors operate at high frequency with low dimensionality, whereas vision are slower and high-dimensional. As a result, fusion pipelines may sacrifice responsiveness~\cite{ichiwara2022contact, zhao2025polytouchrobustmultimodaltactile} or underutilize vision during critical contact stages~\cite{zhao2025touch}. Although \cite{xue2025reactive} introduced an asynchronous module to mitigate these issues, the method has not been validated with high-frequency sensing. Moreover, existing data collection methods are too slow to be precise and often fail to capture robot dynamics in the collected data, which are critical for achieving high precision in dynamic environments. This gap highlights the need for model-based asynchronous fusion strategies that retain the advantages of both vision and force, ensuring neither modality is compromised.

\section{Method}
% compress not mine contribution, list formula not others' core contribution only if is my contribution or need discussion
% intro parts and their usage
% may include why include these simple adaptions
% more analysis of contribution, why and how go beyond baseline(longer than possible in related work)
% skip unecessary parts
% add a subsection of "review of ...(method)" purpose at beginning if possible

% lml equation
% \vspace{-0.15cm}
\subsection{Problem Overview}
% \vspace{-0.1cm}
% task specification
% \yq{Rewrite} 
% We are interested in the installation process for industrial assembly, especially for pushnuts. 
% We divide this process into two stages: an approaching stage, where the robot end-effector is navigated to the installation point using a vision-based method to provide global information; and a contact-rich manipulation stage, where the robot is guided by force-based control to precisely complete the installation.
% Despite recent advances in force-based online learning methods, they still suffer from inefficient adaptation to the environment, and initial exploration in a wrong direction may significantly reduce the overall success rate.
% The major difficulty of this task lies in how to form the contact force during the manipulation stage while referencing to visual information.

The peg insertion process can be divided into two stages, depending on whether contact occurs. 
The first stage is the approaching stage, in which the robot arm is guided to the insertion point using a vision-based method that provides global information. 
The second stage is the contact-rich manipulation stage, where the robot uses force-based control to precisely complete the installation at high frequency while asynchronously utilize visual information. 
The main challenge in this task is adapting to large target velocity and behaving robustly under target motion shifts while maintaining high precision.
We address the challenge by a two stage multi-modality insertion pipeline in which vision dominant approaching is given a larger precision tolerance and residual errors are refined via asynchronous vision–force fusion during the contact-rich insertion stage.

\subsection{Multi-modality Insertion Pipeline}
% \vspace{-0.1cm}
% two stage design, general introduction, and the switch
% notations

% As shown in [fig], our model follows the two-stage scheme for the industrial installation process:
% \begin{enumerate}
%     \item \textbf{Robust Vision based Pose Estimation} This module takes visual input at a certain time step $V_i$ to produce estimated pose $p_i$ for the current time step. The pose $p_i$ serves either as a target for the end-effector to approach during the approaching stage, or a reference for the force-based model to adjust its control output during the manipulation stage, enabling better exploration of the surrounding environment. It consists of two sub-modules: \begin{enumerate}
%         \item %foundation pose
%         Feed-forward pose registration and tracking based on RGB-D input.
%         \item %segmentation to recover from lose tracking
%         A robust RGB-based object tracking method that helps identify occlusions and improves pose estimation accuracy during tracking.
%     \end{enumerate}
%     \item \textbf{Force-based Model with Visual Referencing} We basically follow the design in LML~\cite{tracy2025efficient}. We proposed an additional module that regularize the difference between the expected robot pose and an asynchronously updated target pose from pose estimator. We also implemented failure recovery methods that can recover from vision failure based on force monitoring.
% \end{enumerate}

As shown in Fig.~\ref{fig:workflow}, our model follows a two-stage scheme for peg insertion in a dynamic environment:

\textbf{The approaching stage} moves the robot from a predefined position to roughly contact the moving hole at an unknown position. To balance precision and stability, most existing robot arms employ a PD controller. These controllers perform well when moving toward a distant target given sufficient time. However, when tracking a target with unknown motion at high precision, the robot must react to small per-cycle changes at high frequency. In such cases, existing robot controllers will have a slow response compared to the robot control cycle time. In our experiments, this latency can cause a 1–2 cm mismatch between the robot and a moving target.

To address this issue, we design a cross-coupled PD tracker on top of a position control based robot controller which can accept target velocity as input. Given the current robot pose $\hat{p}_{i-1}$ and the pose estimation result $P_i$ from our vision based pose estimation, adapted from FoundationPose~\cite{wen2024foundationpose} with modifications for dynamic environments. The tracker outputs the expected robot pose $p_i^{expected}$ and velocity $v_i^{expected}$ for robot controller. We update $P_i$ using a Kalman filter to predict the target pose $p_i$ and velocity $v_i$ at the execution time of the current control cycle. We compute $p_i^{expected}$ and $v_i^{expected}$ as:
$$\hat{p}_i^{expected} = p_i + K_{vp}v_i + K_{pp}(p_i - \hat{p}_{i-1})$$
$$v_i^{expected} = v_i + K_{pv}(p_i - \hat{p}_{i-1})$$
This enables the robot to track the hole without large error once a rough alignment in the insertion direction established. We monitor the force reading at every control cycle.

\textbf{The contact-rich manipulation stage} begins once the force sensor detects a contact force exceeding a threshold during the approaching stage. We employ a force-based model with asynchronous visual referencing to reduce residual alignment errors from the approaching stage and to adapt to target motion during insertion. The objective is to maintain contact forces primarily along the insertion direction while minimizing the positional discrepancy between the estimated target pose and the executed robot pose after the current control cycle.

We denote the control $u_i$ for step $i$ as the offset position from current pose $\hat{p}_{i-1}$ to the next pose $\hat{p}_{i}$. We normalize this control to maintain a desired insertion speed along the $z$-axis, which yields a more stable and consistent motion in the intended insertion direction.

\subsection{Force-based Model with Async Visual Referencing}
% \vspace{-0.1cm}
% include simple but necessary adaption(not too long)
% insertion stage, how to recover from not correct insertion initialization failure, how to update LML model to take poses from vision side as reference
% overview

% \yq{Working}

This module primarily operates during the contact-rich manipulation stage, where the estimated base pose is used as a reference input to a force-based decision-making process.
We assume insertion proceeds downward along the $z$-axis, and we define the force along the insertion direction as positive.

\textbf{Force-Based Adaptive Contact Modeling.}
We define the system state $s_i$ as a 19-dimensional vector that concatenates the robot pose, force sensor readings, control input, and a bias term:
$\begin{bmatrix} \hat{p}_{i-1},\; \hat{F}_{i-1},\; u_i,\; 1 \end{bmatrix}^\top$.
We use a linear model $G$, which is widely used for position-force transformation in static environments, to transform the current system state $s_i$ into the predicted force at the next timestep, denoted as $F^{\text{predicted}}_i$.
We initialize the force prediction model as a $6 \times 19$ matrix composed of
a $6 \times 6$ zero matrix for pose,
two $6 \times 6$ identity matrices for force sensor reading and control,
and a $6 \times 1$ zero vector for bias.
This initialization reflects the intuition of bootstrapping with an admittance-like prior.
During manipulation, the robot is governed by a closed-loop controller.

To achieve successful insertion, the model $G$ minimizes contact forces in all directions orthogonal to the $z$-axis.
Along the $z$-axis, positive in the insertion direction, as the contact force during insertion is typically opposite to the motion and therefore negative, we move the robot toward target by subtracting a small value $\delta = 1$ from the current force reading $\hat{F}_{i-1}^z$ until a predefined threshold $F_{\max}$, a negative value because opposite to the insertion direction, is reached.
The expected force after a control cycle is
\begin{equation}
F^{\text{expected}}_i =
\begin{cases}
\mathbf{0}, & \text{orthogonal directions to } z, \\
\max\!\big(\hat{F}_{i-1}^z - \delta,\; F_{\max}\big), & \text{insertion direction } z.
\end{cases}
\label{eq:expected_force}
\end{equation}

In dynamic environments that often exhibit nonlinear behavior and violate quasi-static assumptions, the linear force prediction model $G$ may no longer be valid.
To address this issue, we reformulate nonlinear force behavior into a linearized quasi-static model by introducing state-of-the-art high-frequency online learning for linear models to perform the insertion task~\cite{tracy2025efficient}.
Specifically, the contact model is updated from $G_{i-1}$ to $G_i$ given the current state $s_i$ and the next-step force sensing $\hat{F}_i$ using a Kalman filter.
We discretize the contact-rich manipulation process into small timesteps with a limited motion range and update the transformation matrix at each step as
$G_{i+1} = U(G_i, \hat{F}_i, s_i)$,
where $U$ is the update function defined in~\cite{tracy2025efficient} that updates $G$ based on the robot state $s_i$ and current force/torque sensor readings.
This yields a stepwise linear relationship within each small time or action interval:
$F_{i+1} = G_{i+1} \, s_{i+1}$.
Suppose the underlying environment dynamics evolve only slightly within a single control cycle, the force-prediction model obtained at timestep $i$ remains close to that at timestep $i+1$.
Although each timestep employs a linear model, sequential updates across control cycles allow a globally nonlinear force behavior over longer horizons.

\textbf{Asynchronous Visual Referencing.}
In real-world settings, limited control frequency and slow learning speed, caused by discrepancies between model assumptions and environment dynamics, can prevent online learning from adapting to a moving target or target motion shift quickly enough to eliminate undesired horizontal forces.
These forces may arise from lag between the robot's current pose and the target pose, or from overshoot in the previous motion direction when motion shifts.
Such horizontal forces can further induce large lateral motions, which may violate the assumptions required for linear force modeling.
In addition, force sensing is only reliable within a limited contact range, leaving this range may make force measurements unreliable.
To address these issues, we introduce a visually estimated pose $P$ as a regularizer, transforming the problem from adapting to target motion into making minor adjustments to pose estimation error.
This helps maintain the local linearity assumption and restores reliable force sensing.

Due to high computational latency and the low frequency of sensing and inference, a significant temporal gap exists between image capture time and force-based control reference.
When a new pose estimate $P_j$ with capture time $t_0(j)$ becomes available to the force model, we update a Kalman filter that assumes the target object moves with an unknown constant velocity.
During control prediction, the Kalman filter estimates the target pose $p_i$ and velocity $v_i$, and the terminal velocity of the control cycle is set to $v_i$.

Using the imperfect visual estimation $p_i$ as a prior, we penalize the positional discrepancy between the expected pose after executing the control, $\hat{p}_i^{\text{expected}}$, and $p_i$ via $\|\hat{p}_i^{\text{expected}} - p_i\|_2^2$.
We then formulate the control objective as a weighted sum of force-tracking and pose-regularization terms with weight $\lambda$:
\begin{equation}
\arg \min_{u_i}\;
\big\| F^{\text{expected}}_i - F_i \big\|_2^2
\;+\;
\lambda\, \big\| \hat{p}_i^{\text{expected}} - p_i \big\|_2^2.
\label{eq:control_optimization_deepmind}
\end{equation}

The weight $\lambda$ controls the trade-off between force-based exploration and adherence to the visual prior.
When the target exhibits fast but predictable motion, a larger $\lambda$ constrains exploration and maintains alignment with the estimated pose.
Conversely, when the target is quasi-static or its motion is highly unpredictable, a smaller $\lambda$ places greater emphasis on contact feedback.

In practice, we found that a fixed value of $\lambda = 0.1$ provides robust performance across a wide range of target motions.
Specifically, this setting works well for targets exhibiting approximately linear motion with small variance, as well as trajectories with occasional direction changes along any axis.
Despite variations in target velocity and motion patterns, $\lambda = 0.1$ consistently balances adherence to the visual prior and force-based correction, and therefore used throughout our experiments unless otherwise noted.

\begin{figure}[t]
\centering
\includegraphics[width=\linewidth]{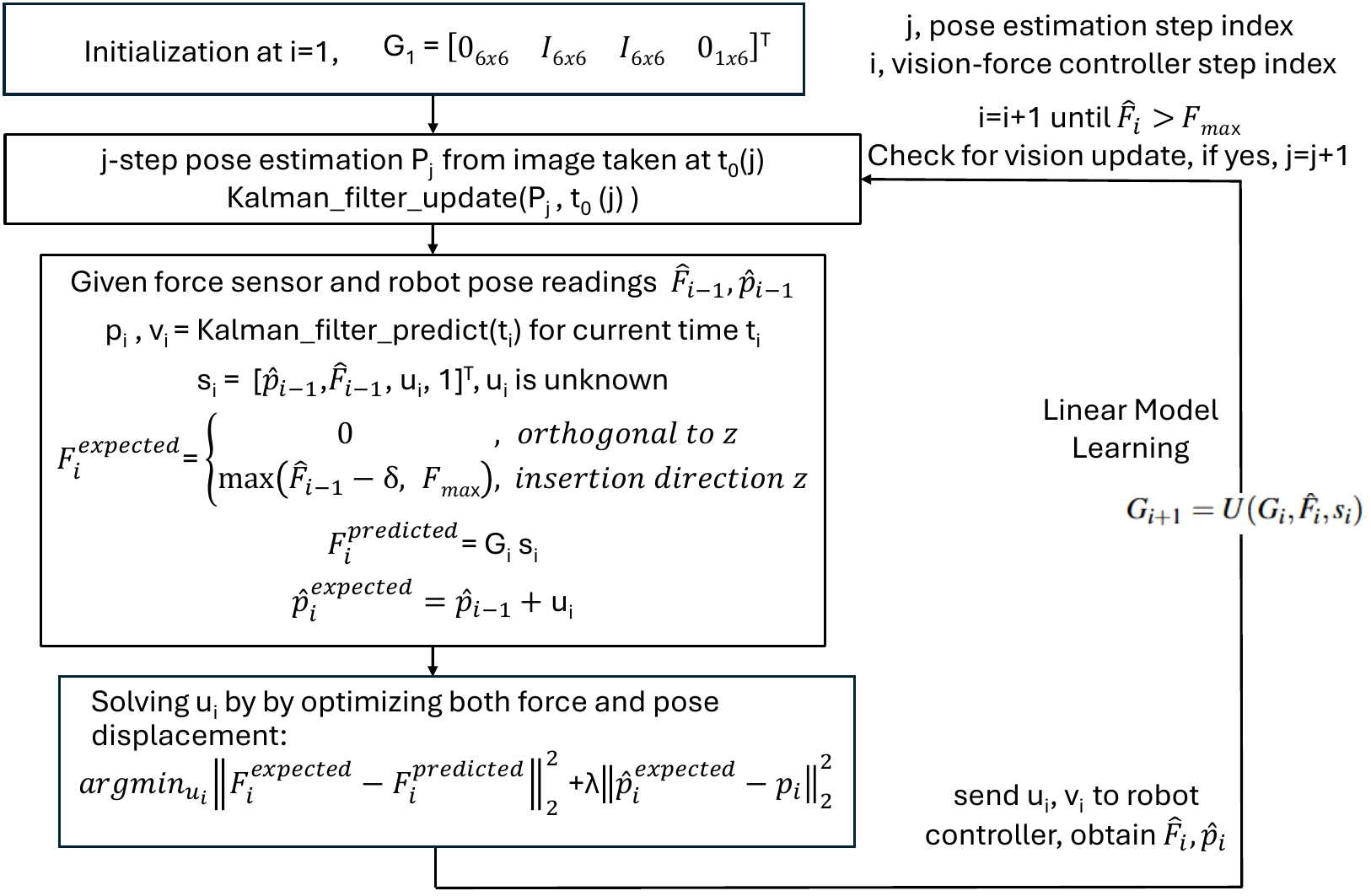}
\captionof{figure}{\textbf{Asynchronous Vision--Force Controller.}
In each control cycle, the force-based model updates itself using force/torque feedback, the robot’s previous control inputs, and its previous state.
When available, visual information is incorporated asynchronously.
The model predicts the next-step control by minimizing the discrepancy between the expected and the sensed force and pose, using the current robot pose, force/torque readings, and visual input.}
\label{fig:force_diag}
\vspace{-0.5cm}
\end{figure}

We mark insertion as complete when the current force reading along the $z$-axis exceeds the threshold $F_{\max}$; the robot then lifts. A diagram of VFAL during the contact-rich manipulation stage is shown in Fig.~\ref{fig:force_diag}.

We also introduce two simple but effective failure recovery methods to enhance robustness in dynamic environments.
The first addresses loss of tracking. If the camera fails to track the object, the robot retreats to a safe position until tracking is restored by the pose estimation module.
The second method addresses inaccurate initial pose during contact-rich insertion. Erroneous or outdated pose estimates, or cases where the robot fails to accurately track the target, may cause the insertion to fail during the approaching stage. These issues are detected by monitoring force peaks that indicates successful fastener engagement. If the force exceeds a threshold $\hat{F}_{\max}$ without a corresponding force peak, the attempt is classified as a failure. In such cases, the robot moves upward and reinitializes the installation process.

\section{Experiment}
% note pushnut
\begin{table*}[t]
\scriptsize
  \centering
      \caption{Success rate and Mean Squared Error (MSE) on full wrench across different models for both fixed-base and mobile-base insertions. We mark the best performed results for each experiment in \textbf{bold}.}
    % \vspace{-0.1cm}
    \resizebox{\textwidth}{!}{\begin{tabular}{|c|c|c|c|c|}
	\hline
	\multirow{1}{*}{\textbf{Description}}& \multirow{1}{*}             {\textbf{Method}}&\multirow{1}{*}                   {\textbf{Modality}}&\multirow{1}{*}{\textbf{Success Rate}}               &\multirow{1}{*}{\textbf{Full Wrench}}                \\
        \hline
         \multirow{6}{*}{\shortstack{1D Mobile Base\\(Cylinder Christmas tree nut)}} & Vision & Vision & 40\%  & 57.11$(\pm 59.65)$  \\

        & Admittance                & Force & 60\% &  70.96$(\pm 29.78)$   \\

         & Online Learning         & Force & 70\%  & 80.95$(\pm 42.56)$  \\

         % & Sync without Kalman                & Force, Vision & 0\% & N/A   \\

         & Synchronous VFAL               & Force, Vision & 90\% & 28.50$(\pm 14.11)$   \\

         % & VFAL without End Velocity (app)         & Force, Vision & 90\%   & 50.89$(\pm 32.91)$  \\

         & VFAL & Force, Vision &  \textbf{100\%} & \textbf{13.95}$(\pm \textbf{5.05})$  \\

        \hline
         \multirow{6}{*}{\shortstack{3D Mobile Base\\(Cross-fin Christmas tree nut)}} & Vision & Vision & 20\%  & 114.60$(\pm 57.30)$  \\

        & Admittance                & Force & 10\% &  106.09$(N/A)$   \\

         & Online Learning         & Force & 0\%  & N/A$(N/A)$  \\

         % & Sync without Kalman                & Force, Vision & 0\% & N/A   \\

         & Synchronous VFAL               & Force, Vision & 60\% & 37.82$(\pm 15.83)$   \\

         % & VFAL without Online Learning (app)         & Force, Vision & 70\%   & 12.11$(\pm 8.04)$  \\

         & VFAL & Force, Vision &  \textbf{80\%} & \textbf{9.22}$(\pm \textbf{5.23})$  \\

        \hline

        % & VFAL ($\lambda=1$) & Force, Vision &  70\% & 73.68$(\pm 50.89)$  \\

        % & VFAL ($\lambda=0.1$) & Force, Vision &  \textbf{80\%} & \textbf{9.22}$(\pm \textbf{5.23})$  \\

        % & VFAL ($\lambda=0.01$) & Force, Vision &  40\% & 67.89$(\pm 68.73)$  \\

        % \hline
  \end{tabular}}
  \label{tab:better_algo_tab}
  \vspace{-0.1cm}
\end{table*}

 We evaluate methods in dynamic environments where pegs are inserted into moving bases with holes. We further evaluate additional experimental settings quantitatively.
More settings are explored through the video demonstrations.

% \vspace{-0.2cm}
\subsection{Experiment Setup}

% \subsubsection{\textbf{Real World Experiment}}
 We use pushnuts, a class of deformable fasteners, as experimental objects.
Specifically, we use Christmas-tree pushnuts (see video demo) with conical, stepped geometries common in industrial and consumer settings.
 These pushnuts are inserted into different bases. All experiments use position control with a capped $z$-axis speed during contact-rich insertion.

We evaluate performance using two metrics. The success rate measures the possibility of successful insertions. A trial is successful if insertion completes with at most one fin outside the hole when the $z$-axis force reaches $F_{\max}$.
We exclude failures occurring during the shared approaching stage before hole contact.
These failures are considered external to contact-rich insertion.
We adapt the full 6D wrench MSE metric from \cite{tong2026scalable} to obtain an aggregate view of adaptation quality, and tailor it to our dynamic insertion task as $\frac{1}{trials \cdot steps}\sum_{trials}\sum_{steps}\left\| \mathbf{m} \odot \mathbf{F}_{observed} \right\|^2, \quad \mathbf{m}=[1,1,0,1,1,1]$. Since the force and torque components are summed in a single squared norm, this quantity serves only as an aggregate indicator of adaptation quality and does not carry a direct physical interpretation. An ablation evaluating the force and torque components separately is provided in Sec.~\ref{sec:sep_ft}.
Only successful trials are included, with 10 trials per method.

\begin{wrapfigure}{r}{0.15\textwidth}
    \captionsetup{type=figure}
    \vspace{-0.9cm}
    \includegraphics[width=\linewidth]{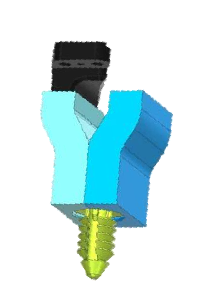}
    \captionof{figure}{\textbf{TPU-based gripper}. It serves as an indicator of force and manipulation status visualizer through its deformation.}
    \label{fig:gripper}
    \vspace{-0.2cm}
\end{wrapfigure}

As shown in Fig.~\ref{fig:gripper}, we design a TPU-based gripper with an internal undercut bowl.
Horizontal force causes visible deformation and eventual ejection, providing clear force and failure indication while protecting the system.

\begin{figure}
    \centering
    \captionsetup{type=figure}
    \includegraphics[width=\linewidth]{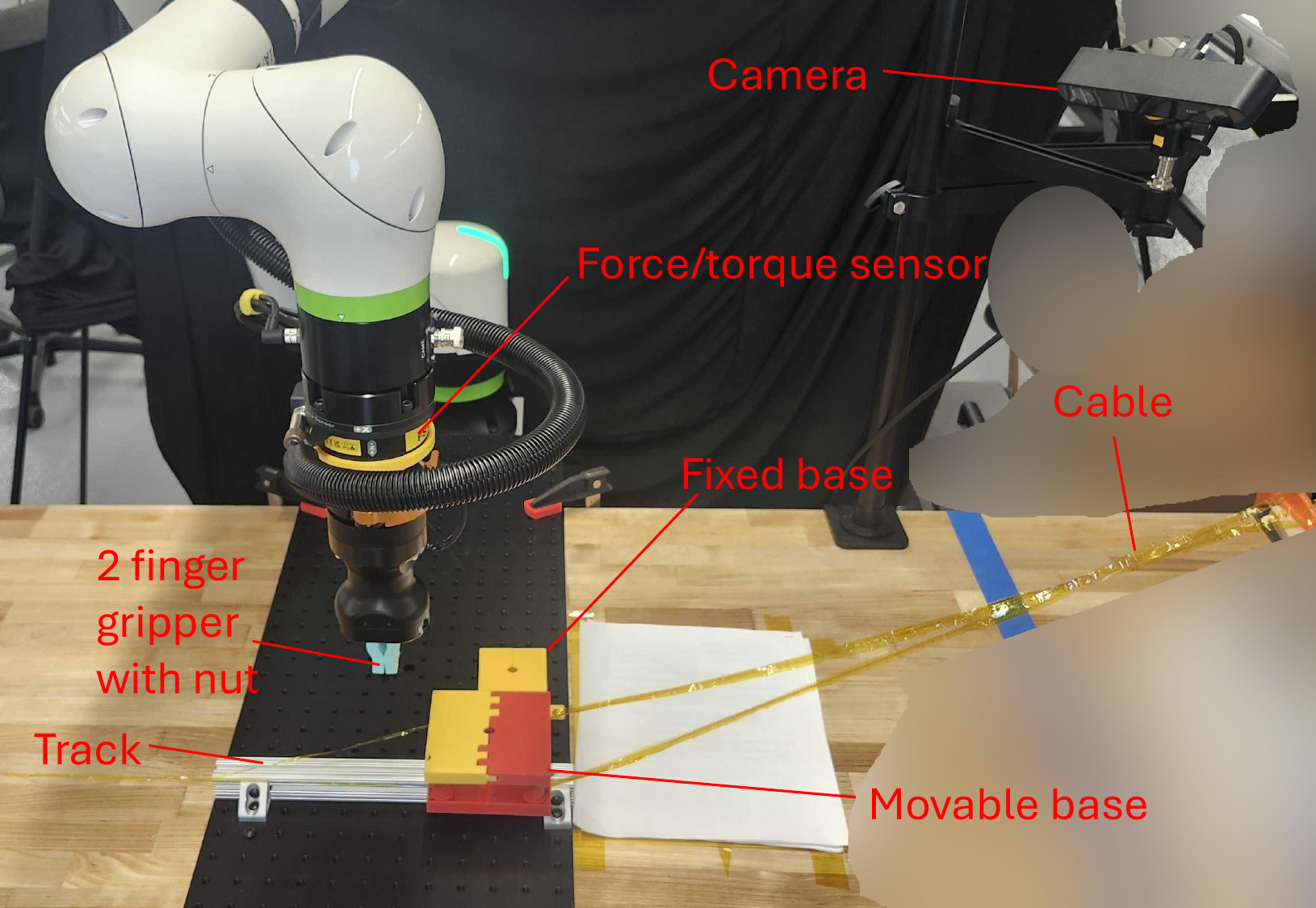}
    \captionof{figure}{\textbf{Real world experiment setup}. In our real world experiment setup, an RGB-D camera is used for pose estimation. The robot, equipped with a force/torque sensor, holds the deformable pushnut and performs the insertion task. The setup includes two types of bases: a fixed base mounted on a motherboard, and a cable-driven movable base mounted on a fixed linear track. This configuration allows us to evaluate insertion performance under both static and dynamic conditions.}
    \label{fig:setup}
    \vspace{-0.6cm}
\end{figure}

The experimental setup is shown in Fig.~\ref{fig:fig1}(a) and Fig.~\ref{fig:setup}.
A table-mounted RGB-D camera is calibrated to the robot using an AprilTag.
AprilTags are used only for camera–robot calibration.
We experiment three bases: fixed, 1D mobile, and 3D mobile.
The 3D mobile base is mounted on a preprogrammed motorized 3D slider. 
A cross-fin pushnut is used for this setting.
The 1D base moves along a linear track via deformable cables.
Only robotic actuation is used for controlled evaluation. 
Track friction and cable elasticity introduce velocity variation under contact.
A cylindrical pushnut is used for 1D experiments. 
All experiments use a FANUC CRX-10iA/L with an FS-15iA force/torque sensor.
Video demo presents and evaluates more different kind of pegs used in the experiment.
Unless noted for synchronous ablations, control and force sensing run at 125 Hz, force data is averaged from a 250 Hz sensor, and vision runs at 20 Hz.

Fixed-base insertions are used to characterize pushnut force patterns. Eight fins in cylinder Christmas tree nut (see Fig.~\ref{fig:fig1}(b)) generate eight force/torque peaks. These signatures enable automatic failure detection and recovery.

% \begin{figure}
%     \centering
%     \captionsetup{type=figure}
%     \includegraphics[width=\linewidth]{figs/force_graph.pdf}
%     \captionof{figure}{\textbf{Force/torque graph} for a fixed-base insertion with one of the pushnuts used in the experiment. Each peak corresponds to a fin engagement during insertion (see video appendix). Due to pose-estimation error during the approaching stage, the two sides did not engage simultaneously, producing two smaller peaks. Force graph for other pegs can be referenced in Appendix.
%     }
%     \label{fig:force}
%     \vspace{-0.6cm}
% \end{figure}

% \vspace{-0.2cm}
\subsection{Quantitative Results and Discussions}

% \begin{figure*}
%     \centering
%     \captionsetup{type=figure}
%     \includegraphics[width=\linewidth]{figs/experiment.pdf}
%     \captionof{figure}{\textbf{Incremental experiment settings for mobile base peg insertion}. We increment from experiment~\ref{it:vision} to experiment~\ref{it:admit} by incorporating force-based control after the approaching stage. Experiment~\ref{it:lml} then introduces online learning. Finally, experiments~\ref{it:sync} and~\ref{it:vfal} fuse vision and force during insertion to explore the performance of fusion methods}
%     \label{fig:experiment}
%     \vspace{-0.5cm}
% \end{figure*}

% \subsubsection{\textbf{Experiment 2}} \label{experiment:dynamic}
We adapt an incremental design for both mobile bases in our experiment.

\begin{enumerate}[(i)]
    \item \label{it:vision} \textbf{Vision}: Vision-only insertion using estimated base pose.
    
    \item \label{it:admit} \textbf{Admittance}: Force-only admittance control under dynamic environment.
    
    \item \label{it:lml} \textbf{Online Learning}: Admittance augmented with online learning from \cite{tracy2025efficient}.

    \item \label{it:sync} \textbf{Synchronous VFAL}: Vision–force fusion with synchronized sensing via blocking.

    % \item \label{it:end} \textbf{VFAL without End Velocity}: This setup follows VFAL, except the velocity at the end of each control cycle is set to the predicted base velocity, as in force-only models. It evaluates the necessity of setting the end velocity of each control cycle from vision.

    \item \label{it:vfal} \textbf{VFAL}: Proposed asynchronous vision-assisted force model learning.
\end{enumerate}

\textbf{Discussion on 1D and 3D environment.} The 1D mobile base is cable-driven and mechanically compliant, allowing it to deform under external forces and thereby tolerate larger positional errors during insertion.
In contrast, the 3D mobile base is motor-driven and significantly stiffer, such that even small pose deviations can induce large contact forces.
Additionally, the cross-fin Christmas-tree pushnut used in the 3D setting has smaller fins and is therefore easier to disengage than the cylindrical pushnut used in the 1D setting. As reflected in Table~\ref{tab:better_algo_tab}, these factors make 3D insertion substantially more demanding, requiring higher manipulation precision and resulting in degraded performance across most metrics.

\textbf{Discussion on vision-only model and force-only models.} Vision-only methods are generally more transferable, as they leverage global context and scene-level information. However, they suffer from higher latency due to sensing and computation overhead, which can cause large errors in dynamic environments. Moreover, visual sensing is typically less precise than force/torque sensing in contact-rich manipulation.
On the 1D mobile base, the system has greater tolerance to detect contact early and adjust the insertion using force feedback. As a result, both force-only methods (\textbf{Admittance} and \textbf{Online Learning}) outperform \textbf{Vision}, as shown in Table~\ref{tab:better_algo_tab}.
In contrast, on the 3D mobile base, the robot has limited tolerance to correct its pose through contact alone. In this setting, vision-only method outperform force-only methods due to their ability to provide more robust global information.

\textbf{Discussion on vision-force fusion.} Results on the mobile base experiments show that combining vision and force (\textbf{Synchronous VFAL} and \textbf{VFAL}) enables higher performance than either modality alone (\textbf{Vision}, \textbf{Admittance}, and \textbf{Online Learning}). Vision–force fusion benefits from precise local force feedback while retaining global visual context, leading to improved robustness and accuracy in dynamic insertion tasks.

\textbf{Discussion on asynchronous.} The comparison between \textbf{Synchronous VFAL} and \textbf{VFAL} further demonstrates that incorporating visual information asynchronously improves performance. By asynchronous fusion of vision into force, the controller achieves lower latency from higher control frequency, which directly translates into better task performance.

% \vspace{-0.2cm}
\subsection{Ablation Studies}

\subsubsection{Ablation on Online Learning}

To evaluate the effect of online learning, we conduct experiments on the 3D mobile base and compare \textbf{VFAL} with \textbf{VFAL without Online Learning}, as shown in Table~\ref{tab:online}.

\vspace{-0.1cm}
\begin{table}[h]
\centering
\caption{Ablation on Online Learning (3D Mobile Base)}
\vspace{-0.2cm}
\begin{tabular}{|c|c|c|}
\hline
	\multirow{1}{*}             {\textbf{Method}}&\multirow{1}{*}      {\textbf{Success Rate}}               &\multirow{1}{*}{\textbf{Full Wrench MSE}}                \\
        \hline

         VFAL without Online Learning & 70\%   & 12.11$(\pm 8.04)$  \\

         VFAL &  \textbf{80\%} & \textbf{9.22}$(\pm \textbf{5.23})$  \\
         
\hline

\end{tabular} \label{tab:online}
\vspace{-0.2cm}
\end{table}

The results indicate that online learning improves both the success rate and full wrench MSE. Online learning is still preferred even if asynchronous fusion of vision and force is provided.

\subsubsection{Ablation on $\lambda$ Selection}

As described in the method section, we set the visual prior regularization weight to $\lambda = 0.1$ in all main experiments. To validate this choice, we evaluate different values of $\lambda$ on the 3D mobile base. Each setting is tested over 10 trials.

\vspace{-0.1cm}
\begin{table}[h]
\centering
\caption{Ablation on $\lambda$ Selection (3D Mobile Base)}
\vspace{-0.2cm}
\begin{tabular}{|c|c|c|}
\hline
	\multirow{1}{*}             {\textbf{Method}}&\multirow{1}{*}      {\textbf{Success Rate}}               &\multirow{1}{*}{\textbf{Full Wrench MSE}}                \\
        \hline

        $\lambda=1$ &  70\% & 73.68$(\pm 50.89)$  \\

        $\lambda=0.1$ &  \textbf{80\%} & \textbf{9.22}$(\pm \textbf{5.23})$  \\

        $\lambda=0.01$ &  40\% & 67.89$(\pm 68.73)$  \\
         
\hline
\end{tabular}
\end{table}
\vspace{-0.2cm}

Among the tested values, \textbf{$\lambda = 0.1$} achieves the highest success rate and the lowest force/torque error. Increasing $\lambda$ to 1 degrades performance, as an overly strong visual prior suppresses reliable force feedback, which is inherently more precise during contact. When $\lambda$ is reduced to $0.01$, visual guidance becomes too weak, resulting in force-dominated behavior and substantially lower performance.
Nevertheless, even at \textbf{$\lambda = 0.01$}, the method still outperforms pure force-only baselines on the 3D mobile base.

\subsubsection{Ablation on the robot controller}

We conduct an ablation study on a high frequency robot controller while keeping the force sensing frequency matching the pose tracking frequency as in \textbf{Synchronous VFAL}. This experiment evaluates whether the asynchronous vision force fusion design is the main source of VFAL's performance gain over \textbf{Synchronous VFAL}.

\vspace{-0.1cm}
\begin{table}[h]
\centering
\caption{Ablation on the robot controller (3D Mobile Base)}
\vspace{-0.2cm}
\begin{tabular}{|c|c|c|}
\hline
	\multirow{1}{*}             {\textbf{Method}}&\multirow{1}{*}      {\textbf{Success Rate}}               &\multirow{1}{*}{\textbf{Full Wrench MSE}}                \\
        \hline

        Synchronous VFAL &  60\% & 37.82$(\pm 15.83)$  \\

        High Frequency Controller &  60\% & 54.95$(\pm 41.11)$  \\

        VFAL &  \textbf{80\%} & \textbf{9.22}$(\pm \textbf{5.23})$  \\
         
\hline
\end{tabular}
\end{table}
\vspace{-0.2cm}

The performance of \textbf{High Frequency Controller} matches \textbf{Synchronous VFAL}, while \textbf{VFAL} reaches a much higher performance. Raising the controller rate is not the main source of VFAL's performance gain.

\subsubsection{Ablation on force/torque}
\label{sec:sep_ft}

We report force and torque MSE separately, alongside the full wrench MSE, to characterize contact behavior in more detail.

\vspace{-0.1cm}
\begin{table}[h]
\centering
\caption{Ablation on force/torque (3D Mobile Base)}
\vspace{-0.2cm}
\begin{tabular}{|c|c|c|c|}
\hline
	\multirow{1}{*}             {\textbf{Method}}&\multirow{1}{*}      {\textbf{Force}}               &\multirow{1}{*}{\textbf{Torque}}      &\multirow{1}{*}{\textbf{Full Wrench}}          \\
        \hline

        Vision & 100.32$(\pm 51.39)$ & 14.28$(\pm 5.91)$ & 114.60$(\pm 57.30)$ \\
        
        Admittance & 95.02$(N/A)$ & 11.07$(N/A)$ & 106.09$(N/A)$ \\
        
        Online Learning & N/A$(N/A)$ & N/A$(N/A)$ & N/A$(N/A)$ \\
        
        Sync VFAL & 33.26$(\pm 13.99)$ & 4.56$(\pm 1.94)$ & 37.82$(\pm 15.83)$ \\
        
        VFAL & \textbf{8.07}$(\pm \textbf{4.56})$ & \textbf{1.15}$(\pm \textbf{0.68})$ & \textbf{9.22}$(\pm \textbf{5.23})$ \\
                 
\hline
\end{tabular}
\end{table}
\vspace{-0.2cm}

\textbf{VFAL} still performs the best among all evaluated methods under both the force and torque MSE metrics.

\subsubsection{Ablation on full system}

We additionally report the approaching stage and the end-to-end success rate alongside the insertion results.

\vspace{-0.1cm}
\begin{table}[h]
\centering
\caption{Ablation on full system (3D Mobile Base)}
\vspace{-0.2cm}
\begin{tabular}{|c|c|c|c|}
\hline
	\multirow{1}{*}             {\textbf{Method}}&\multirow{1}{*}      {\textbf{Approaching}}               &\multirow{1}{*}{\textbf{Insertion}}      &\multirow{1}{*}{\textbf{Overall}}          \\
        \hline

        Vision & 10/17  & 2/10 & 2/17 \\

        Admittance  & \textbf{10/10} &  1/10 & 1/10   \\

         Online Learning & 10/12  & 0/10 & 0/12  \\

         Synchronous VFAL  & 10/13 & 6/10 & 6/13   \\

         VFAL &  10/12 & \textbf{8/10} & \textbf{8/12}  \\
         
\hline
\end{tabular}
\end{table}
\vspace{-0.2cm}

The total number of trials for each method varies, since the main contribution of this work concerns the insertion stage. Thus, we keep collecting data until 10 trials successfully enter the insertion stage. The same results are also visualized as a Sankey diagram in the real world robot demo, experiment 1.

% \vspace{-0.2cm}
\subsection{Real World Robot Demo}
% experiment setup for each demo
% \yq{If these are all quantitative results. Add sentences saying we show the results in demo video or supp.}
We deploy our method on a FANUC CRX-10iA/L robot arm to qualitatively demonstrate insertion behaviors in both static and dynamic scenarios, including different pegs, different version of target motion, and failure recovery. Demonstration videos are provided in the supplementary material.

\begin{enumerate}[(i)]
    \item \textbf{Experiment 1:} Peg insertion on a 3D mobile base under the same setup as the 3D mobile base experiments. Results for all trials are visualized at the beginning.

    \item \textbf{Experiment 2:} Insertion of different peg types, including a pushnut and a rigid cylinder peg, into a 3D mobile base.
    
    \item \textbf{Experiment 3$\And$4:} Peg insertion on a 1D mobile base under 1D mobile base experiment speed (Experiment 3) and doubled speed (Experiment 4).
    
    \item \textbf{Experiment 5:}  Recovery from unintended panel collisions caused by tracking failures in the approach stage.
    
    \item \textbf{Experiment 6:} Recovery from occlusion or losing tracking, preventing potential collisions.

    \item \textbf{Experiment 7:} Peg insertion into a fixed base mounted on a table to present ideal force behavior for insertion.

    \item \textbf{Experiment 8:} Peg insertion into a base with unknown and noisy human-driven motion in a 2D plane.
\end{enumerate}

\section{Conclusion and Limitation}
% summarize, potential future direction
In this work, we develop a modality fusion framework that asynchronously combines vision and force sensor readings without reducing sensor frequency by introducing a regularization term between the estimated pose and the expected pose after motion, and by setting a target speed after each control cycle. This framework enables the robot to perform precise peg insertion into a movable hole at a high control frequency. We evaluate our framework in the real world. Our model achieves high success rate, low force error, and demonstrates strong adaptability to dynamic environments.

The proposed framework has limitation. It assumes that the peg has already been grasped. In real-world scenarios, however, the robot must first perform reliable grasping. 

%%%%%%%%%%%%%%%%%%%%%%%%%%%%%%%%%%%%%%%%%%%%%%%%%%%%%%%%%%%%%%%%%%%%%%%%%%%%%%%%

{\small
\bibliographystyle{IEEEtranN}
\balance
\bibliography{sections/reference}
}

\newpage

\end{document}